# A Novel Feature Extraction for Robust EMG Pattern Recognition

Angkoon Phinyomark, Chusak Limsakul, and Pornchai Phukpattaranont

**Abstract**—Varieties of noises are major problem in recognition of Electromyography (EMG) signal. Hence, methods to remove noise become most significant in EMG signal analysis. White Gaussian noise (WGN) is used to represent interference in this paper. Generally, WGN is difficult to be removed using typical filtering and solutions to remove WGN are limited. In addition, noise removal is an important step before performing feature extraction, which is used in EMG-based recognition. This research is aimed to present a novel feature that tolerate with WGN. As a result, noise removal algorithm is not needed. Two novel mean and median frequencies (MMNF and MMDF) are presented for robust feature extraction. Sixteen existing features and two novelties are evaluated in a noisy environment. WGN with various signal-to-noise ratios (SNRs), i.e. 20-0 dB, was added to the original EMG signal. The results showed that MMNF performed very well especially in weak EMG signal compared with others. The error of MMNF in weak EMG signal with very high noise, 0 dB SNR, is about 5-10% and closed by MMDF and Histogram, whereas the error of other features is more than 20%. While in strong EMG signal, the error of MMNF is better than those from other features. Moreover, the combination of MMNF, Histrogram of EMG and Willison amplitude is used as feature vector in classification task. The experimental result shows the better recognition result in noisy environment than other success feature candidates. From the above results demonstrate that MMNF can be used for new robust feature extraction.

**Index Terms**—Electromyography (EMG), Feature extraction, Pattern recognition, Robustness, Man-machine interfaces.

—————————— ♦ ——————————

## 1 INTRODUCTION

SURFACE Electromyography (sEMG) signal is one of the electrophysiological signals, which is extensively studied and applied in clinic and engineering. In this research, the application of sEMG signal in assistive technology and rehabilitation engineering is paid attention. Main application of these fields is the control of the prosthesis or other assistive devices using the different patterns of sEMG signal [1-2]. Nevertheless, the major drawback of EMG pattern recognition is the poor recognition results under conditions of existing noises especially when the frequency characteristic of noise is random. Major types of noise, artefact and interference in recorded sEMG signal are electrode noise, electrode and cable motion artifact, alternating current power line interference, and other noise sources such as a broad band noise from electronic instrument [3-4]. The first three types of noise can be removed using typical filtering procedures such as band-pass filter, band-stop filter, or the use of well electrode and instrument [3-4] but the interferences of random noise that fall in EMG dominant frequency energy is difficult to be removed using previous procedures. Generally, white Gaussian noise (WGN) is used to represent the random noise in sEMG signal analysis [5-6]. Adaptive filter or wavelet denoising algorithm, advance digital signal filter, has been received considerable attention in the removal of WGN [7-8]. However, WGN cannot be removed one hundred percent and sometimes some important part of sEMG signals are removed with noise even if we use adaptive filter and wavelet denoising algorithm. The broad band and random frequency characteristic of noise in this group is a main reason that make it difficult to be removed. Moreover, the amplitude of noise is bigger than the sEMG signal amplitude; the amplitude of raw signal is about 50 μV-100 mV [9].

In EMG-based pattern recognition, sEMG signal is preprocessed the spectral frequency component of the signal and extracted some features before performing classification [1]. Normally, in preprocessing and signal condition procedure, method to remove noise is a significant step to reduce noises and improve some spectral component part [3]. Next important step, feature extraction, is used for highlighting the relevant structures in the sEMG signal and rejecting noise and unimportant sEMG signal [5]. The success of EMG pattern recognition depends on the selection of features that represent raw sEMG signal for classification. This study is motivated by the fact that the limitation of the solutions to remove WGN in the preprocessing step and EMG-based gestures classification need to do the extraction step. The selection of feature that torelance of WGN and the modified of existing EMG feature to improve the robust property are proposed. As a result, WGN removal algorithms in the preprocessing step are not needed.

————————————————

- *Angkoon Phinyomark is with Department of Electrical Engineering, Prince of Songkla University, Songkhla, Thailand 90112.*
- *Chusak Limsakul is with the Department of Electrical Engineering, Prince of Songkla University, Songkhla, Thailand 90112.*
- *Pornchai Phukpattaranont is with the Department of Electrical Engineering, Prince of Songkla University, Songkhla, Thailand 90112.*

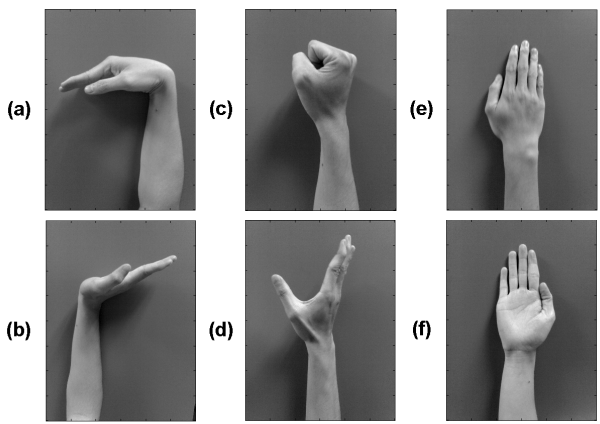

Fig. 1. Estimated six upper limb motions (a) wrist flexion (b) wrist extension (c) hand close (d) hand open (e) forearm pronation (d) forearm supination.

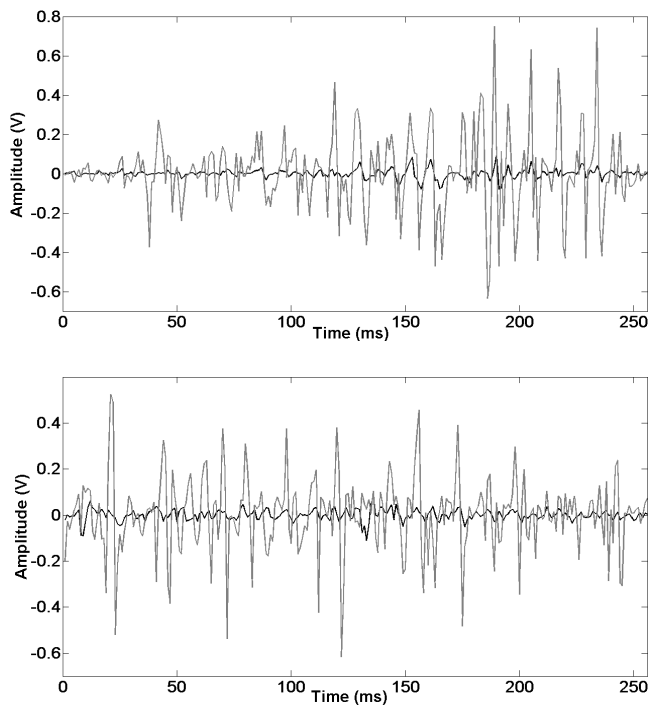

Fig. 2. Strong sEMG signal (gray line) and weak sEMG signal (black line) of (a) wrist extension motion. (b) hand open motion.

From the literatures, the development of robust feature extractions in speech, texture, and image are presented [10-11] but there is no selection and modification of robust EMG feature extraction. There are some evaluations about the effect of noise with EMG features [5, 12-14]. However, these literatures attend to the quality of EMG features in maximum class separability point of view. The description and discussion about the robustness are inferiority. Furthermore, features that used to evaluate in the literatures are not fair with the available methods today. In 1995, Zardoshti-Kermani et al. [5] evaluated seven features in time domain and frequency domain. WGN with 0 to 50% of rms amplitude signal are used to test the effect of noise. The cluster separability index and classification result are presented that histrogram of EMG is the better feature in very high noise (50% of rms amplitude signal). Later, in 2003, thirteen features with combination and various orders are tested the robustness property by Boostani et al. [12]. One level, one tenth of sEMG peak-to-peak amplitude, of 50 Hz interference and random noise is considered and the sensitivity of feature is reported. In addition, our previous work [13-14] compared the effect of eight features and their relevant features with 50 Hz interference and WGN. The results of mean square error (*MSE*) criterion show that Willison amplitude with 5 mV threshold parameter is the best feature compared to the other features.

However, there is an increase in EMG feature methods that is published in many literatures this day. In this paper, sixteen features in time domain and frequency domain from the literatures [5, 12-17] are used to test the robustness with the additive WGN at various signal-to-noise ratios (*SNR*s). Moreover, the effect of the level of signal amplitude was tested. Eighteen features that used in this research represent most features in EMG pattern recognition. Generally, most of the attempts to extract features from sEMG signal can be classified into three categories including time domain, frequency domain, and time-frequency domain [1]. We considered only former two categories because they have computational simplicity and they have been widely used in research and in clinical practice. In addition, two novel feature calculations using frequency properties are presented. We modified the mean frequency and median frequency by calculating the mean and median of amplitude spectrum instead of power spectrum that we called Modifed Mean Frequency (MMNF) and Modified Median Frequency (MMDF). This paper is organized as follows. Experiments and data acquisition are presented in Section 2. Section 3 presents a description of EMG feature extraction methods in time domain and frequency domain. In addition, the evaluation criterion is introduced. Results and discussion are reported in Section 4, and finally the conclusion is drawn in Section 5.

## 2 EXPERIMENTS AND DATA ACQUISITION

In this section, we depict our experimental procedure for recording sEMG signals. The sEMG signal was recorded from flexor carpi radialis and extensor carpi radialis longus of a healthy male by two pairs of Ag-AgCl Red Dot surface electrodes on the right forearm. Each electrode was separated from the other by 2 cm. A band-pass filter of 10-500 Hz bandwidth and an amplifier with 60 dB gain was used. Sampling frequency was set at 1 kHz using a 16 bit analog-to-digital converter board (National Instruments, DAQCard-6024E).

A volunteer performed four upper limb motions including hand open, hand close, wrist extension, and wrist flexion as shown in Fig. 1 (a-d). In this study, the effect of signal strength was performed by divided the sEMG signal to two types: strong sEMG signal and weak sEMG signal. Strong sEMG signals were collected from extensor carpi radialis longus in hand close and wrist flexion and were collected from flexor carpi radialis in hand open and wrist extension as shown in Fig. 2 (gray line). In addition, the others motion and electrode channel are weak sEMG signals as shown in Fig. 2 (black line). Ten datasets were collected for each motion. The sample size of the sEMG signals is 256 ms for the real-time constraint that the response time should be less than 300 ms. This dataset was used for the *MSE* criterion that represent the effect of noise with the value of EMG features.

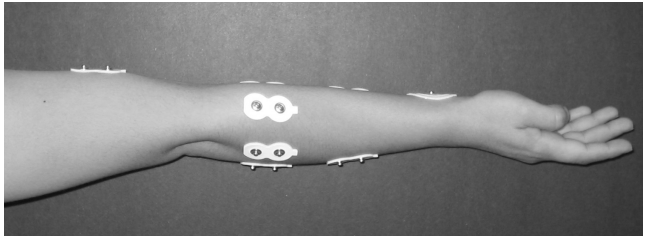

Fig. 3. The eight electrode placements of the right forearm.

The second dataset is used to evaluate the performance of the classification results of EMG features in noisy environment. Seven upper limb motions including hand open, hand close, wrist extension, wrist flexion, forearm pronation, forearm supination and resting as shown in Fig. 1 and eight electrode positions on the right forearm as shown in Fig. 3 were used in the classification procedure to measure the performance of the EMG feature space quality with WGN. This dataset was acquired by the Carleton University in Canada [17]. A duo-trode Ag-AgCl surface electrode (Myotronics, 6140) was used and an Ag-AgCl Red-Dot surface electrode (3M, 2237) was placed on the wrist to provide a common ground reference. This system set a bandpass filter with a 1–1000 Hz bandwidth and amplifier with a 60 dB (Model 15, Grass Telefactor). The sEMG signals were sampled by using an analog-to-digital converter board (National Instruments, PCI-6071E), and the sampling frequency was 3 kHz. However, in pattern recognition, downsample of EMG data from 3 kHz to 1 kHz was done. Each trial of the database consisted of four repetitions of each motion. There are six trials in each subject. Three subjects were selected in this study. More details of experimentals and data acquisition are described in [17].

## 3 METHODOLOGY

Eighteen time domain features and frequency domain features are described in this section. Thirteen time domain variables are measured as a function of time. Because of their computational simplicity, time domain features or linear techniques are the most popular in EMG pattern recognition. Integrated EMG, Mean absolute value, Modified mean absolute value 1, Modified mean absolute value 2, Mean absolute value slope, Simple square integral, Variance of EMG, Root mean square, Waveform length, Zero crossing, Slope sign change, Willison amplitude, and Histogram of EMG are used to test the performance. All of them can be done in real-time and electronically and it is simple for implementation. Features in this group are normally used for onset detection, muscle contraction and muscle activity detection. Moreover, features in frequency domain are used to represent the detect muscle fatigue and neural abnormalities, and sometime are used in EMG pattern recognition. Three traditional and two modified features in frequency spectrum are performed namely autoregressive coefficients, mean and median frequencies, modified mean and median frequencies. Afterward, the evaluation methods of two criterions that used to measure the robustness property of the whole features are introduced.

### 3.1 Time Domain Feature Extraction

#### 3.1.1 Integrated EMG

Integrated EMG (IEMG) is calculated as the summation of the absolute values of the sEMG signal amplitude. Generally, IEMG is used as an onset index to detect the muscle activity that used to oncoming the control command of assistive control device. It is related to the sEMG signal sequence firing point, which can be expressed as

$$\text{IEMG} = \sum_{n=1}^{N} |x_n|, \qquad (1)$$

where $N$ denotes the length of the signal and $x_n$ represents the sEMG signal in a segment.

#### 3.1.2 Mean Absolute Value

Mean Absolute Value (MAV) is similar to average rectified value (ARV). It can be calculated using the moving average of full-wave rectified EMG. In other words, it is calculated by taking the average of the absolute value of sEMG signal. It is an easy way for detection of muscle contraction levels and it is a popular feature used in myoelectric control application. It is defined as

$$\text{MAV} = \frac{1}{N} \sum_{n=1}^{N} |x_n|. \qquad (2)$$

#### 3.1.3 Modified Mean Absolute Value 1

Modified Mean Absolute Value 1 (MMAV1) is an extension of MAV using weighting window function $w_n$. It is shown as

$$\text{MMAV1} = \frac{1}{N} \sum_{n=1}^{N} w_n |x_n|,$$

$$w_n = \begin{cases} 1, & \text{if } 0.25N \leq n \leq 0.75N \\ 0.5, & \text{otherwise} \end{cases}. \qquad (3)$$

#### 3.1.4 Modified Mean Absolute Value 2

Modified Mean Absolute Value 2 (MMAV2) is similar to MMAV1. However, the smooth window is improved in this method using continuous weighting window function $w_n$. It is given by

$$\text{MMAV2} = \frac{1}{N} \sum_{n=1}^{N} w_n |x_n|, \qquad (4)$$

$$w_n = \begin{cases} 1, & \text{if } 0.25N \leq n \leq 0.75N \\ 4n/N, & \text{if } 0.25N > n \\ 4(n-N)/N, & \text{if } 0.75N < n \end{cases}$$

### 3.1.5 Mean Absolute Value Slope

Mean Absolute Value Slope (MAVSLP) is a modified version of MAV. The differences between the MAVs of adjacent segments are determined. The equation can be defined as

$$\text{MAVSLP}_i = \text{MAV}_{i+1} - \text{MAV}_i . \quad (5)$$

### 3.1.6 Simple Square Integral

Simple Square Integral (SSI) uses the energy of the sEMG signal as a feature. It can be expressed as

$$\text{SSI} = \sum_{n=1}^{N} |x_n|^2 . \quad (6)$$

### 3.1.7 Variance of EMG

Variance of EMG (VAR) uses the power of the sEMG signal as a feature. Generally, the variance is the mean value of the square of the deviation of that variable. However, mean of EMG signal is close to zero. In consequence, variance of EMG can be calculated by

$$\text{VAR} = \frac{1}{N-1} \sum_{n=1}^{N} x_n^2 . \quad (7)$$

### 3.1.8 Root Mean Square

Root Mean Square (RMS) is modeled as amplitude modulated Gaussian random process whose RMS is related to the constant force and non-fatiguing contraction. It relates to standard deviation, which can be expressed as

$$\text{RMS} = \sqrt{\frac{1}{N} \sum_{n=1}^{N} x_n^2} . \quad (8)$$

The comparison between RMS and MAV feature is reported in the literatures [3, 18]. Clancy et al. experimentally found that the processing of MAV feature is equal to or better in theory and experiment than RMS processing. Furthermore, the measured index of power property that remained in RMS feature is more advantage than MAV feature.

### 3.1.9 Waveform Length

Waveform length (WL) is the cumulative length of the waveform over the time segment. WL is related to the waveform amplitude, frequency and time. It is given by

$$WL = \sum_{n=1}^{N-1} |x_{n+1} - x_n| . \quad (9)$$

All of these features above, 3.1.1-3.1.9, are computed based on sEMG signal amplitude. From the experimental results, the pattern of these features is similar. Hence, we selected the robust feature representing for the other features in this group. The results and discussion is presented in Section 4.1.

### 3.1.10 Zero Crossing

Zero crossing (ZC) is the number of times that the amplitude value of sEMG signal crosses the zero y-axis. In EMG feature, the threshold condition is used to abstain from the background noise. This feature provides an approximate estimation of frequency domain properties. It can be formulated as

$$ZC = \sum_{n=1}^{N-1} \left[ \text{sgn}(x_n \times x_{n+1}) \cap |x_n - x_{n+1}| \geq \text{threshold} \right];$$

$$\text{sgn}(x) = \begin{cases} 1, & \text{if } x \geq \text{threshold} \\ 0, & \text{otherwise} \end{cases} . \quad (10)$$

### 3.1.11 Slope Sign Change

Slope Sign Change (SSC) is similar to ZC. It is another method to represent the frequency information of sEMG signal. The number of changes between positive and negative slope among three consecutive segments are performed with the threshold function for avoiding the interference in sEMG signal. The calculation is defined as

$$SSC = \sum_{n=2}^{N-1} \left[ f\left[ (x_n - x_{n-1}) \times (x_n - x_{n+1}) \right] \right];$$

$$f(x) = \begin{cases} 1, & \text{if } x \geq \text{threshold} \\ 0, & \text{otherwise} \end{cases} . \quad (11)$$

### 3.1.12 Willison Amplitude

Willison amplitude (WAMP) is the number of times that the difference between sEMG signal amplitude among two adjacent segments that exceeds a predefined threshold to reduce noise effects same as ZC and SSC. The definition is as

$$\text{WAMP} = \sum_{n=1}^{N-1} f(|x_n - x_{n+1}|);$$

$$f(x) = \begin{cases} 1, & \text{if } x \geq \text{threshold} \\ 0, & \text{otherwise} \end{cases} . \quad (12)$$

WAMP is related to the firing of motor unit action potentials (MUAP) and the muscle contraction level.

The suitable value of threshold parameter of features in ZC, SSC, and WAMP is normally chosen between 10 and 100 mV that is dependent on the setting of gain value of instrument. Nevertheless, the optimal threshold that suitable for robustness in sEMG signal analysis is evaluated and discussed in Section 4.1.

### 3.1.13 Histogram of EMG

Histogram of EMG (HEMG) divides the elements in sEMG signal into $b$ equally spaced segments and returns the number of elements in each segment. HEMG is an extension version of the ZC and WAMP features. The effect of various segments is tested and expressed in Section 4.1.

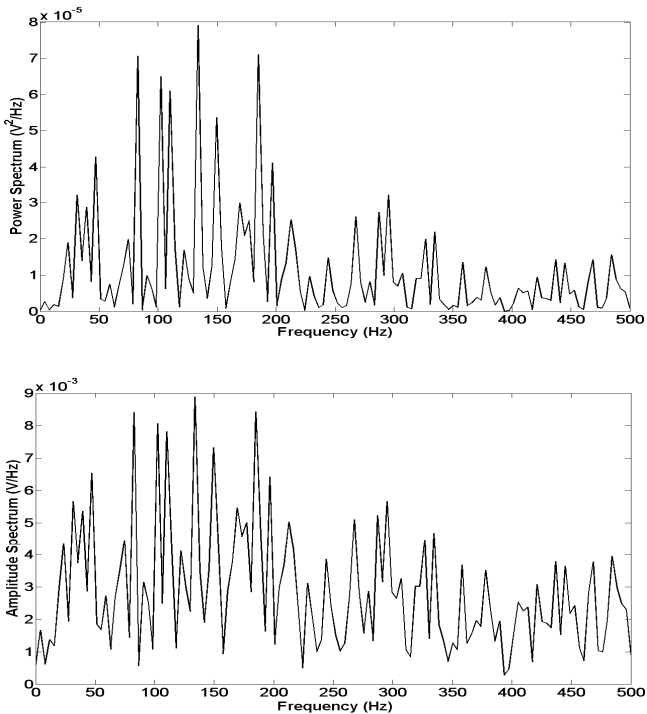

Fig. 4. Power spectrum (×10⁻⁵) (top) and amplitude spectrum (×10⁻³) (bottom) of noisy sEMG signal at 20 dB *SNR* in hand close motion.

### 3.2 Frequency Domain Feature Extraction

#### 3.2.1 Autoregressive Coefficients

Autoregressive (AR) model described each sample of sEMG signal as a linear combination of previous samples plus a white noise error term. AR coefficients are used as features in EMG pattern recognition. The model is basically of the following form:

$$x_n = -\sum_{i=1}^{p} a_i x_{n-i} + w_n , \qquad (13)$$

where $x_n$ is a sample of the model signal, $a_i$ is AR coefficients, $w_n$ is white noise or error sequence, and $p$ is the order of AR model.

The forth order AR was suggested from the previous research [19]. However, the orders of AR between the first order and the tenth order are found. The results are discussed in Section 4.1.

#### 3.2.2 Modifed Median Frequency

Modified Median Frequency (MMDF) is the frequency at which the spectrum is divided into two regions with equal amplitude. It can be expressed as

$$\sum_{j=1}^{\text{MMDF}} A_j = \sum_{j=\text{MMDF}}^{M} A_j = \frac{1}{2}\sum_{j=1}^{M} A_j , \qquad (14)$$

where $A_j$ is the sEMG amplitude spectrum at frequency bin $j$.

#### 3.2.3 Modifed Mean Frequency

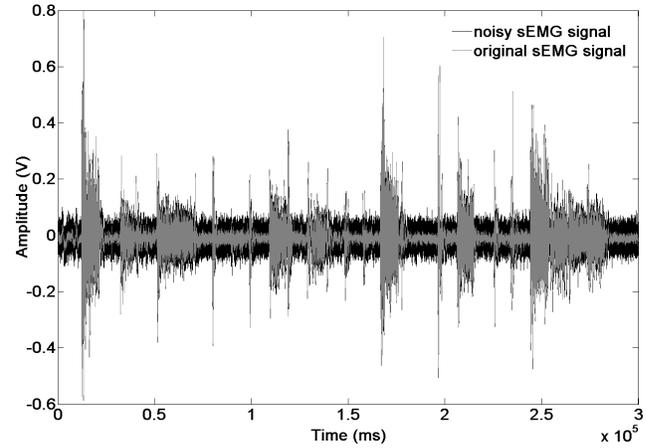

Fig. 5. Original sEMG (gray line) signal and noisy sEMG signal at 5 dB *SNR* (black line) in six upper limb motions.

Modified Mean Frequency (MMNF) is the average frequency. MMNF is calculated as the sum of the product of the amplitude spectrum and the frequency, divided by the total sum of spectrum intensity, as in

$$\text{MMNF} = \sum_{j=1}^{M} f_j A_j \Big/ \sum_{j=1}^{M} A_j , \qquad (15)$$

where $f_j$ is the frequency of spectrum at frequency bin $j$.

#### 3.2.4-3.2.5 Mean Frequency and Median Frequency

Traditional median frequency (MDF) and traditional mean frequency (MNF) are calculated based on power spectrum. We can calculate using the sEMG power spectrum $P_j$ instead of amplitude spectrum $A_j$. They can be expressed as

$$\sum_{j=1}^{\text{MDF}} P_j = \sum_{j=\text{MDF}}^{M} P_j = \frac{1}{2}\sum_{j=1}^{M} P_j , \qquad (16)$$

$$\text{MNF} = \sum_{j=1}^{M} f_j P_j \Big/ \sum_{j=1}^{M} P_j . \qquad (17)$$

The outline of amplitude spectrum and power spectrum is similar but the amplitude value of amplitude spectrum is larger than amplitude value of power spectrum as shown in Fig. 4. Moreover, the variation of amplitude spectrum is less than the power spectrum. For that reason, variation of MMNF and MMDF is also less than traditional MNF and MDF.

### 3.3 Evaluation methods

The percentage error (*PE*) is used to evaluate the quality of the robust of WGN of EMG features, as in

$$PE = \left| \frac{feature_{clean} - feature_{noise}}{feature_{clean}} \right| \times 100\% , \qquad (18)$$

where *feature_clean* denotes the feature vector of the original sEMG signal and *feature_noise* represents the feature vector of the noisy sEMG signal. WGN at different level is added

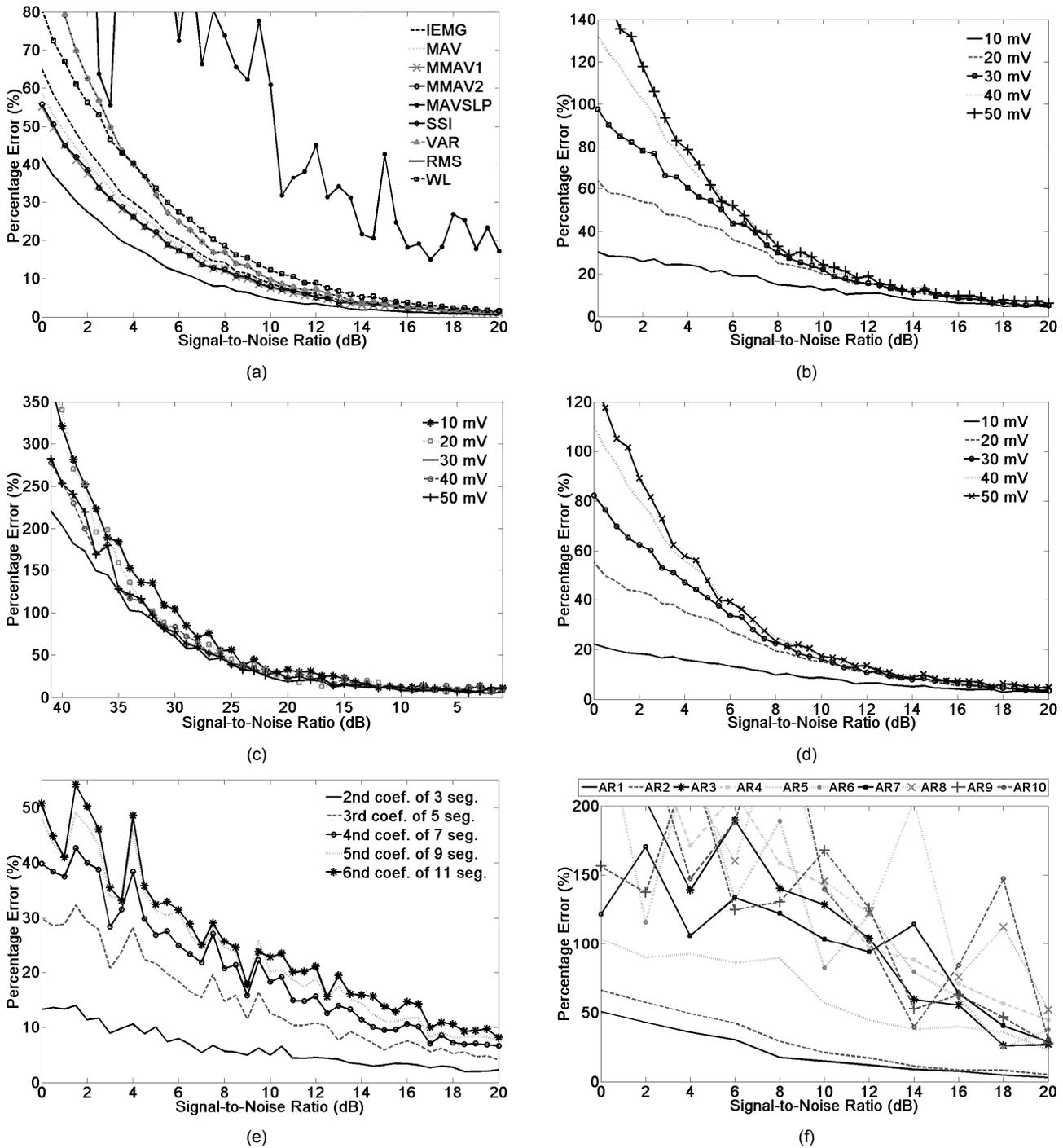

Fig. 6. Average *PE* of sEMG signals of (a) features in time domain based on signal amplitude, 3.1.1-3.1.9 (b) ZC with various threshold value (10-50 mV) (c) SSC with various threshold value (10-50 mV) (d) WAMP with various threshold value (10-50 mV) (e) HEMG with various segment parameters (3, 5, 7, 9, 11 segments) (f) AR coefficients with various orders (1-10) at various signal-to-noise ratios (20-0 dB) in four motions.

to the original sEMG signal.

The performance of the methods is the best when *PE* is the smallest value. We calculated average *PE* for each motion with ten repeated datasets. Therefore, there are 80 datasets with four motions and two channels for each feature and noise level was varied from 20 to 0 dB *SNR* for each dataset. Moreover, WGN was added 10 times in each noise level to confirm the results. *SNR* is calculated by

$$SNR = 10\log\frac{P_{clean}}{P_{noise}}, \quad (19)$$

where $P_{clean}$ is power of the original sEMG signal and $P_{noise}$ is power of WGN.

The classification rate (*CR*) is used to evaluate the quality of the recognition system with the noisy environment of sEMG signal. The performance of the methods is the best when the *CR* values still have the same value with the noisy sEMG signals. Original sEMG signal and noisy sEMG signal were sent to hand movement recognition. In this study, we evaluated the performance of robust features in pattern recognition view point with Myoelectric Control development (MEC) toolbox [17]. The window size is 256 ms and window slide is 64 ms for the real-time constraint that the response time should be less than 300 ms. The feature vector of selected robust features was evaluated by linear discriminant anslysis classifier (LDA) and majority vote (MV) post-processing was performed in this study. In summary, the robust features should have the small value of *PE* and still have maximum classification accuracy.

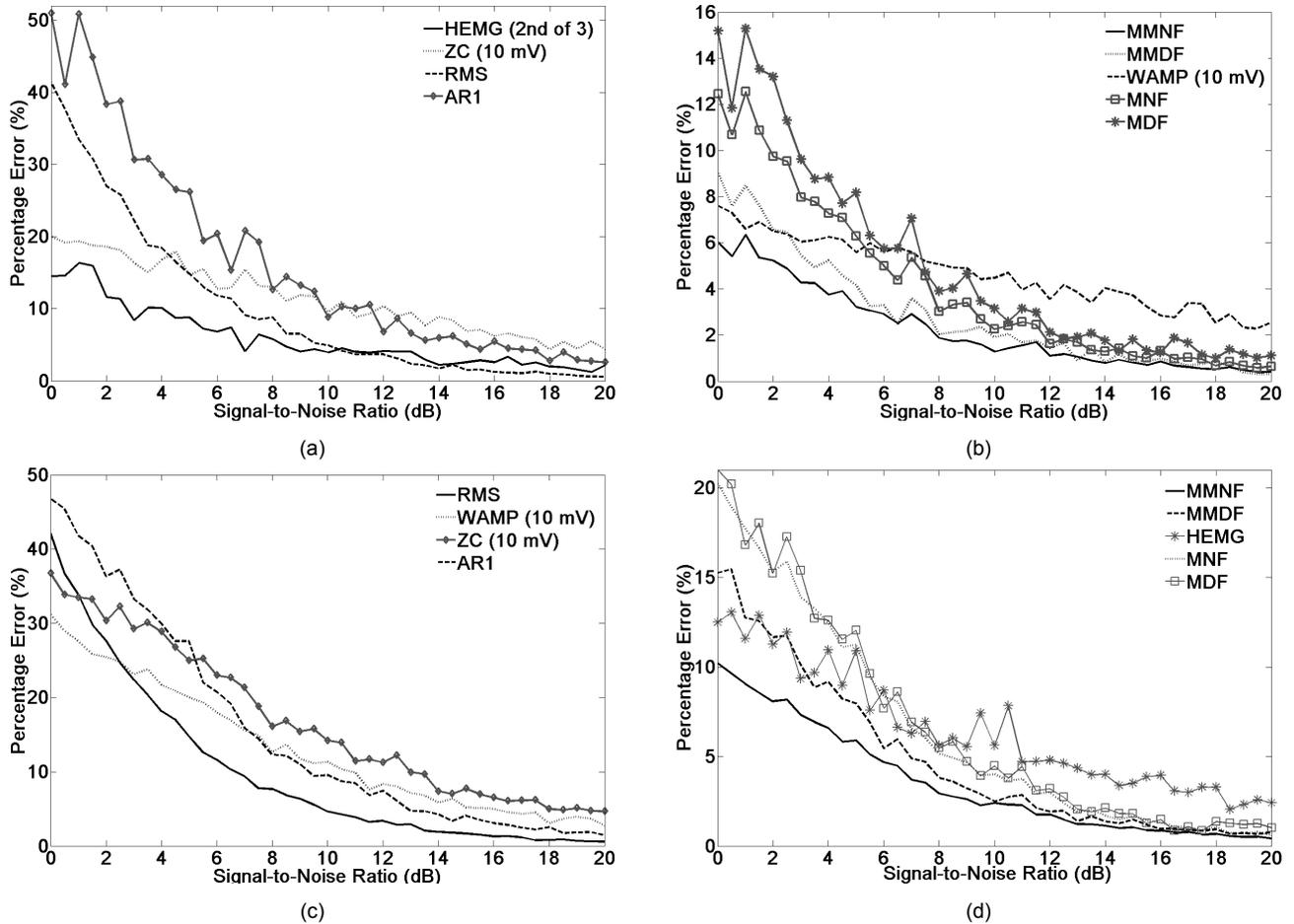

Fig. 7. (a-b) Average *PE* of strong sEMG signals of nine selected features at various signal-to-noise ratios (20-0 dB *SNR*s) in four motions. (c-d) Average *PE* of weak sEMG signals of nine selected features at various signal-to-noise ratios (20-0 dB *SNR*s) in four motions.

## 4 RESULTS AND DISCUSSION

### 4.1 The Quality of the Robustness of EMG Features with WGN

To test the robustness of sixteen traditional features and two novel features, we measured the *PE* with sEMG signal with additive WGN. The results of *PE* are plotted for *SNR*s from 20 dB to 0 dB, as shown in Fig. 6-7, in practice; we can select feature extraction to be suitable for each application depend on the level of interference of sEMG system. For the easy way to describe the results of a large number of features, we discussed and evaluated the features that have the same pattern in recognition and evaluated some parameter of each feature in Fig. 6. As a result, only nine representatives are discussed as the results shown in Fig. 7.

In Fig. 6 (a), the *PE* of time domain features computed using sEMG signal amplitude demonstrates that RMS results in powerful performance in robust noise tolerance than the other features. Hence, RMS feature is used to represent the features in this group. Fig. 6 (b-d) present the evaluation of suitable value of threshold. Threshold value was chosen between 10 and 50 mV. The optimal threshold is 10 mV for ZC and WAMP but the suitable threshold of SSC is 30 mV. However, the minimum PE of SSC is higher than ZC and WAMP. ZC and WAMP with 10 mV threshold value are selected for the representative features of this group. Afterward, the second bin of the third segment HEMG was adopted from the result in Fig. 6 (e) and the first-order of AR is chosen because the PE of the other AR orders are much bigger than the first one as shown in Fig. 6 (f).

Therefore, we evaluated the performance of robustness of nine representative features namely RMS, ZC and WAMP with 10 mV threshold, HEMG with 2nd bin, AR order 1, MNF, MDF, MMNF, and MMDF. Two types of sEMG signal, strong signals and weak signals, are used to evaluate the robust of nine features. The weak sEMG signal has the effect of interference more than the strong sEMG signal. In practice, we can select the robust features to be suitable for each application. Fig. 7 (a) and Fig. 7 (b) show the average *PE* of strong sEMG. For strong sEMG signals and low noise, *SNR* more than 10 dB, MMNF has the smallest average *PE*, followed closely by the MMDF, MNF, and MND. For *SNR* less than 10 dB that showed high noise, the *PE* of MNF and MDF rapidly increased and *SNR* less than 3 dB that showed very high noise, WAMP has the average *PE* close to MMNF. The average error of MMNF in strong sEMG with very high noise, 0 dB *SNR*, is only 6%. Moreover in wrist extension and hand open from extensor carpi radialis longus, it is only 3.5%. HEMG and ZC have slightly larger error compared to the first group in Fig. 7 (b). The *PE* of RMS and AR1 are large that they were expected to perform poorly.

TABLE 1
CLASSIFICATION RATE (%) OF 7 EMG FEATURES
USING LEAVE-ONE-OUT VALIDATION

| Method | Level of SNR noise | | | |
|---|---|---|---|---|
| | Clean | 20 dB | 15 dB | 10 dB |
| HEMG | 60.7835 | 49.1590 | 41.7926 | 34.6817 |
| WL | 79.3059 | 34.1707 | 14.4347 | 12.5186 |
| WAMP | 86.6298 | 92.2504 | 47.0087 | 21.6095 |
| MMNF | 41.1326 | 36.3636 | 32.6804 | 17.1386 |
| MAV,WL, ZC,SSC | 95.6781 | 67.4260 | 22.5676 | 7.9838 |
| RMS,AR2 | 96.4871 | 89.8872 | 64.8712 | 25.2714 |
| HEMG,WAMP,MMNF | 93.0807 | 96.1891 | 64.0622 | 28.1243 |

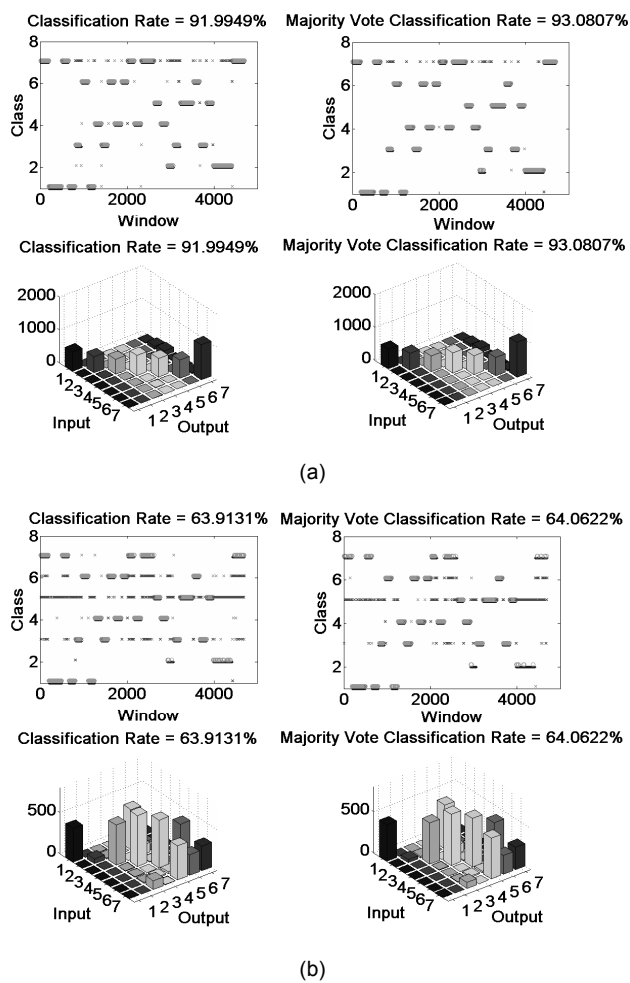

Fig. 8. Plot of the classification results as a function of time (Top) and plot of confusion matrix (Below) of our multi-feature set with (a) original sEMG signal. (b) noisy sEMG signal, 15 dB *SNR*.

The average *PE* of weak sEMG signals shown in Fig. 7 (c-d) clearly demonstrate that MMNF is the best robustness feature and closed by MMDF and HEMG, whereas the error of other features is more than 20%. In very high noise, 0 dB *SNR*, it provides average *PE* about 10% and the *PE* of wrist extension from flexor carpi radialis is only 5%. Other feature results are similar to the results of strong sEMG signal but the results of *PE* of weak sEMG signal are larger than the *PE* of strong sEMG signal. The results above show that MMNF was the best feature comparing with others in four motions. In summary, MMDF and WAMP can be used with HEMG for multi-feature. Hence, it is compared the classification results in noisy environment with other successful individual feature and multi-feature sets from the literatures [5, 14-15, 20] in Section 4.2.

### 4.2 The Quality of the Recognition System of EMG Features with the Noisy Environment

Four individual features and three multi-feature sets are examined in this study. The classification results of seven representative features are reported in Table 1. Leave-one-out validation was used to guarantee an exact performance measure for this dataset. The first single robust feature is HEMG that suggested by Zardoshti-Kermani et al. [5] and is comfirmed with our result in Section 4.1. The second feature, WL is recommended by Oskoei and H. Hu [15] that it is the best single feature in their experiment. Lastly, two individual features, WAMP and MMNF, are aprroved by our previous experiment [14] and the experimental results in this paper. The recognition results of single feature did not perform well but it is commonly used in EMG recognition system. The *CR* of WAMP in clean and low noisy environment is good but its *CR* is rapidly decreased in high noise. The *CR* of HEMG is still stability even if noise increases. In addition, no surprising that the CR of WL in noisy environment is poor that comfired by the result in Section 4.1 and the CR of MMNF is poor because of the limitation of their ability to discriminate between classes. However, in practice, we are usually combined this feature with other features to get the useful information features. Because of only one feature per channel of feature that provided from features in time domain and frequency domain, it is effective and small enough to combine with other features for a more powerful feature vector and avoiding additional significant computational burden.

Therefore, the multi-features are the excellent way to provide the powerful performance in recognition system. The combination of robust features namely, MMNF, WAMP, and HEMG is compared with two successful and popular multi-features that was used by Hudgins et al. [20], MAV, WL, ZC and SSC, and was recommended by Oskoei and H. Hu [15] , consits of RMS and AR2. From the experiment in Table 1, our robust multi-features group provides more excellent discriminatory power for a classifier than Hudgins's and Oskoei's multi-features group in noisy environment. Moreover, the observation from the classification results as a function of time and plot of confusion matrix of our multi-feature set with orginal sEMG signal and noisy sEMG signal are shown in Fig. 8.

## 5 CONCLUSION

The objectives of this study were to present a novel feature that tolerate with white Gaussian noise. Sixteen traditional features and two novel features in time domain and frequency domain were tested. Results showed that a modified mean frequency (MMNF) is the best feature comparing with others in the quality of the robustness of EMG features with WGN point of view. MMNF has average error only 6% in strong sEMG signals and 10% in weak sEMG signal at *SNR* value of 0 dB and MMNF has average error only 0.4% in both strong and weak sEMG signals at *SNR* value of 20 dB. In addition, MMNF and other robust features (WAMP and HEMG) are used as an input to the EMG pattern recognition. The experiment shows that these features are the excellent candidates for a multi-source feature vector. From the above experiment results, it is shown that MMNF can be used as feature in augmenting the other features for a more powerful robust feature vector. Future work is recommended to combine the new multi-feature sets with MMNF to be tested in other classifer types.

## ACKNOWLEDGMENT

This work was supported in part by the Thailand Research Fund through the Royal Golden Jubilee Ph.D. Program (Grant No. PHD/0110/2550), and in part by NECTEC-PSU center of excellence for rehabilitation engineering and Faculty of Engineering, Prince of Songkla University. The authors would like to acknowledge the support of Dr. A.D.C. Chan, from the Carleton University for providing the datasets.

**Angkoon Phinyomark** received the B.Eng. degree in computer engineering with first class honors in 2008 from Prince of Songkla University, Songkhla, Thailand, where he is currently working toward the Ph.D. degree in electrical engineering. His research interests are primarily in the area of biomedical signal processing and rehabilitation engineering. He is a student member of the IEEE.

**Chusak Limsakul** received the B.Eng. degree in electrical engineering from King Mongkut's Institute of Technology Ladkrabang Bangkok, Thailand in 1978 and D.E.A and D. Ing. from Institute National des Sciences Appliquees de Toulouse, France in 1982 and 1985, respectively. He started working as lecturer at the department of electrical engineering at Prince of Songkla University in 1978. He is currently an associate professor of electrical engineering and vice president for research and graduate studies at Princeof Songkla University. His research interests are biomedical signal processing, biomedical instrumentation and neural network.

**Pornchai Phukpattaranont** was born in Songkla, Thailand. He received the B. Eng. and M. Eng. degrees in electrical engineering from Prince of Songkla University in 1993 and 1997, respectively, the Ph.D. degree in electrical engineering from the University of Minnesota, in 2004. He is currently an assistant professor of electrical engineering at Prince of Songkla University. His research interests are ultrasound contrast imaging, ultrasound signal processing, medical image processing, and biomedical signal processing. Dr.Phukpattaranont is a member of the IEEE.